\documentclass[transmag]{IEEEtran}
\pdfoutput=1
\usepackage{latexsym}
\usepackage{graphicx}
\usepackage{amsfonts,amssymb,amsmath}
\usepackage{hyperref}
\usepackage{xcolor}
\usepackage{subcaption}
\usepackage{booktabs} 
\usepackage[linesnumbered,ruled,vlined]{algorithm2e} 
\usepackage{algpseudocode} 
\usepackage{graphicx}
\usepackage{subcaption}
\usepackage{tabularx}
\usepackage{epsfig}  
\usepackage{xcolor}
\usepackage{amsthm}
\usepackage{multirow}
\usepackage{amssymb}
\usepackage{xcolor}
\usepackage{pdfpages}
\usepackage{url}
\usepackage{fancyhdr}
\usepackage{soul}
\usepackage{bbm}

\DeclareMathOperator*{\argmax}{arg\,max}

\newif\ifrev

\revfalse

\ifrev
  \newcommand{\rev}[1]{{\hl{#1}}}
\else
  \newcommand{\rev}[1]{\textcolor{black}{#1}} 
\fi

\begin{document}

\title{Deployment Optimization for Shared e-Mobility Systems with Multi-agent Deep Neural Search}

\author{%
  Man Luo\textsuperscript{1,2},  Bowen Du\textsuperscript{2}, Konstantin Klemmer\textsuperscript{2},  Hongming Zhu\textsuperscript{3} and Hongkai Wen\textsuperscript{1,2*}\thanks{* Corresponding author.}   \\
  \textsuperscript{1} The Alan Turing Institute, UK \hspace*{10pt} 
  \textsuperscript{2} University of Warwick, UK  \hspace*{10pt}
  \textsuperscript{3} Tongji University, China \\
  
}

\IEEEtitleabstractindextext{

\begin{abstract}
Shared e-mobility services have been widely tested and piloted in cities across the globe, and already woven into the fabric of modern urban planning. This paper studies a practical yet important problem in those systems: how to deploy and manage their infrastructure across space and time, so that the services are \textit{ubiquitous} to the users while \textit{sustainable} in profitability. However, in real-world systems evaluating the performance of different deployment strategies and then finding the optimal plan is prohibitively expensive, as it is often infeasible to conduct many iterations of trial-and-error. We tackle this by designing a high-fidelity simulation environment, which abstracts the key operation details of the shared e-mobility systems at fine-granularity, and is calibrated using data collected from the real-world. This allows us to try out arbitrary deployment plans to learn the optimal given specific context, before actually implementing any in the real-world systems. In particular, we propose a novel multi-agent neural search approach, in which we design a hierarchical controller to produce tentative deployment plans. \rev{The generated deployment plans are then tested using a multi-simulation paradigm, i.e., evaluated in parallel, where the results are used to train the controller with deep reinforcement learning.} With this closed loop, the controller can be steered to have higher probability of generating better deployment plans in future iterations. The proposed approach has been evaluated extensively in our simulation environment, and experimental results show that it outperforms baselines e.g., human knowledge, and state-of-the-art heuristic-based optimization approaches in both service coverage and net revenue.
\end{abstract}


}

\maketitle


\section{Introduction}
\label{sec:intro}
Shared electric mobility (e-mobility) systems are becoming ubiquitous and forming a considerable part of our transportation paradigm in urban environments. By integrating them with other modalities, cities could optimize commuting experience, mitigate congestion and elevate air quality, thereby improving living standards. Therefore, deploying more such services is bound to offer an improved convenience and raise acceptance levels among users, thus boosting utilization and profitability over the long run. In this paper, we study the problem of \textit{deployment optimization} for shared e-mobility systems, which aims to find the optimal way of deploying the shared e-mobility infrastructure (stations in our case) during operation, achieving the desired balance between gains and cost. 

This has become increasingly important as the services mature: they have now passed the initial fast growing stage in which high deployment cost can be tolerated, but are more keen to fine-tune themselves to increase profit margin, while improving coverage and service level if possible. To achieve that, the shared e-mobility systems need to carefully select which stations to open or close and when, so that the overall performance is optimize throughout lifespan. Specifically, this particular problem falls into a broader class of optimization tasks in the context of e-mobility systems, which have attracted attentions from various communities. For instance, the work in~\cite{Du:KDD:2018} addresses a similar problem of charger planning for electric vehicle (EV) sharing services. It proposes a demand-aware planning approach and uses heuristics to approximate the optimal plan which ensures both pervasive coverage (in terms of reaching as many POIs as possible) and satisfies sufficient user charging demand. However, this approach only performs one-off optimization relying on aggregated historical demand estimates, which won't be able to adapt as situation changes. On the other hand, the recent work in~\cite{Liu:SIGSPATIAL:2019} considers the incremental cases, but essentially it uses greedy-based approaches to re-compute for charger planning, which may not be efficient in many cases. In addition, most of the existing work aims to maximizing certain objectives given fixed budgets, such as service coverage and satisfied demand.

\rev{We consider a more realistic case, where the deployment of the shared e-mobility system could be optimized dynamically as they operate, with budget that could change over time. Concretely, we would like to find the dynamic deployment plan that guides system deployment through time, which maximizes given objectives (e.g., service coverage or revenue) while satisfing the constraints on budget. In particular, we consider the budget for infrastructure deployment as a function of time, which could depend on the expected income generated by the potential deployment plan, i.e., the revenue of the system under that plan should at least not exceed the cost of adjusting its stations (the system is self-sustaining). In reality, comparing to fixed budgets, such strategies could better cater for varying user demand and provide substantial improvement in performance, e.g., we could temporally close stations that are often quiet on specific days (weekdays vs. weekends), while deploying ``overflow'' stations in the presence of demand surges.} However, it is not trivial to find the optimal deployment plans, in that i) for such mobility systems operating at city scales, the search space of station deployment can be prohibitively large; and ii) for a given deployment plan, accurately evaluating its performance could be challenging. This may be easy for static metrics such as service coverage, but for those that involve interactions between users and the systems, e.g., revenue of orders, it is often difficult to evaluate without actually running the systems with the deployment plan. Using historical data to extrapolate is one option (as in~\cite{Du:KDD:2018} and~\cite{Liu:SIGSPATIAL:2019}), but as we show later the performance can be very limited.

\begin{figure*}[ht!]
\centering
\includegraphics[width=0.8\textwidth]{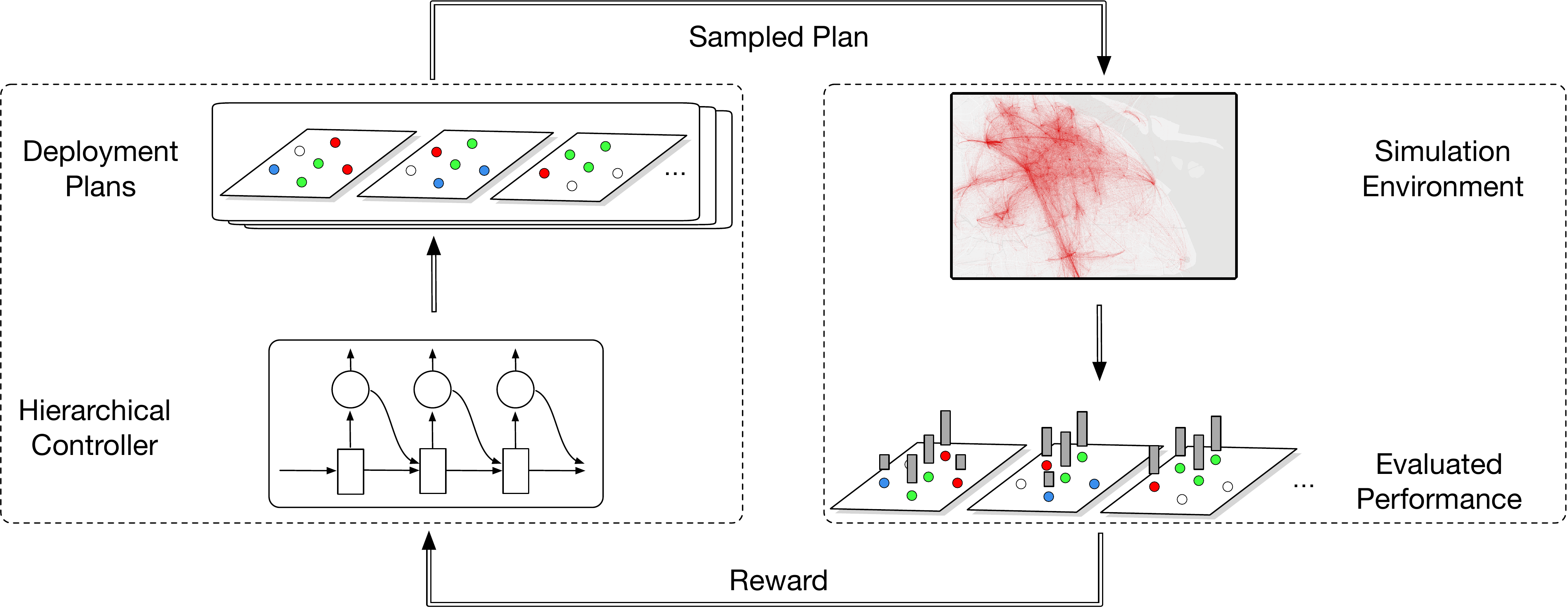}
\caption{Workflow of the proposed hierarchical neural search approach. The hierarchical controller generates possible deployment plans, which are then run in the simulation environment to evaluate their performance. The reward signals (e.g., net revenue and service coverage) obtained are used to update the controller parameters with reinforcement learning.}
\label{fig:arch}
\end{figure*}

To address the challenges, in this paper we build a high-fidelity simulation environment for shared e-mobility systems, which is able to capture the operation details of such systems at fine-granularity. The simulation environment is calibrated with data collected from a real-world shared e-mobility system for a year, to ensure that it can faithfully simulate the real system behaviors and user interactions. With the simulation environment, we design a novel multi-agent neural search algorithm to address the deployment optimization problem. Specifically, we consider a hierarchical controller, which iteratively proposes deployment plans, and evaluates their performance. The results are propagated back to the controller as rewards, whose parameters are updated accordingly so it can generate better plans in the future, as shown in Fig.~\ref{fig:arch}. Concretely, the technical contributions of this paper are as follows:
\begin{itemize}
    
    \item \rev{We design a high-fidelity simulation environment for shared e-mobility systems at city scales. Our simulation environment abstracts the key functionalities that are crucial in practice, while is able to capture the fine-grained details of operations in the shared e-mobility system. We calibrate the simulation with multi-modal real world data, which provides support for training and testing the proposed algorithms.} 
    
    \item We propose a novel multi-agent neural search algorithm to address the deployment optimization problem. We formulate the problem with multi-agent reinforcement learning (MARL) framework, and develop a novel hierarchical controller architecture, which learns to search for optimal deployment plans efficiently. 
    
    \item We evaluate the proposed approach extensively, and show that development plans proposed by our approach significantly outperform the actual plan used in the real-world, and the state-of-the-art optimization approaches, achieving improvements in both revenue and coverage. 
    
\end{itemize}



\section{Related Work}
\label{sec:related}

\noindent \textbf{Shared Mobility Services. }
Recently there is a surging research interests in shared mobility systems in cities, e.g., from understanding taxi demand~\cite{Tong:KDD:2017} to reposition shared bikes~\cite{lowalekar:AAAI:2017} and optimising shared mobility systems to achieving the desired balance between profit and fairness~\cite{nanda:AAAI:2020}. EV-sharing in particular, as one of the environmental friendly services, has attracted extensive attention from various angles and incubated many interesting problems, such as energy consumption estimation~\cite{Pamula:MDPI:2020}, charging scheduling~\cite{Wang:RTSS:2018, Yuan:ICDCS:2019} and infrastructure planning~\cite{Sarker:IMWUT:2018}. These studies mainly consider public transportation, e.g., electric taxis or buses, while our work focuses on privately-owned EV sharing services, which is fundamentally different. For instance, for electric taxis their networks of stations~\cite{Wang:MobiCom:2019} are mainly used for charging, whose service coverage depends on individual taxi drivers. However in our case, the users can only access (renting, returning and charging) vehicles at the available stations, and thus the deployment of stations will have much more direct impact on the entire system. Moreover, comparing to traditional vehicle sharing systems, optimizing the operation of EV sharing services is more complex due to the unique properties of EVs. 

\noindent \rev{\textbf{Urban Mobility Systems Simulation. }}
\rev{Building simulation environments to study urban mobility systems has become a popular approach in existing work~\mbox{\cite{soriguera2018simulation, maciejewski2013simulation, seow2009collaborative}}. However, most of those solutions rely on oversimplified assumptions, while only considering the historical data of the mobility system for training and testing. For instance, the simulator in~{\cite{soriguera2018simulation}} first simulates potential users randomly across space, and then those generated users need to keep ``walking'' to different stations until an available bike is found, which is not realistic nor efficient. The simulation environment for taxi in~{\cite{seow2009collaborative}} assumes the taxis need to move randomly on the road network, while in~{\cite{maciejewski2013simulation}} the potential users have to wait until they are picked up by an arriving taxi, both of which are often not true in practice. In the context of simulating facility/infrastructure deployment, most of the existing work only considers one-time~{\cite{Du:KDD:2018}}, or multiple-stages~{\cite{Liu:SIGSPATIAL:2019}} of infrastructure deployment or expansion. Our simulation environment, on the other hand, simulates the system operation and the actual interactions between users and the system with much finer details, and supports dynamic/continuous infrastructure deployment, i.e., stations can be opened or shut down at arbitrary times.}

\noindent \textbf{Deep Learning in Mobility Applications. }
Deep learning techniques have been used in various mobility applications due to their superior performance, e.g., improving service levels of taxi~\cite{Wei:IEEEACESS:2017}, predicting user demand~\cite{Luo:IMWUT:2020}, and reposition bikes or vehicles~\cite{Pan:AAAI:2019}. \rev{In particular, deep reinforcement learning has been introduced to solve various challenging mobility problems, such as road management~{\cite{pan2020road}}, order dispatching~{\cite{Li:WWW:2019}}, and rebalancing~{\cite{Luo:ijcai:2020}}. Due to their distributed nature, many of those mobility applications can be modeled as multi-agent games, which can be well solved by deep reinforcement learning. For instance, the work in~{\cite{li:Kdd:2018}} designs a spatio-temporal reinforcement learning approach to dynamically reposition bikes in the bike-sharing system. The work in~{\cite{lin:Kdd:2018}} proposes a multi-agent reinforcement learning framework to tackle the fleet management problem, while~{\cite{Li:WWW:2019}} addresses the order dispatching problem for ride sharing systems using mean field approximation. It has been shown that in those applications, deep reinforcement learning often achieves better performance, e.g., in terms of reducing potential customer loss, or increasing the gross merchandise values, than the traditional rule-based or optimization based approaches, especially when the problem structure is complex. }

\noindent \textbf{\rev{Infrastructure Deployment and Optimisation.}}
\rev{Our work is also related to the problem of infrastructure deployment and optimisation, which has been extensively investigated in various application scenarios, such as determining locations for automotive services~{\cite{wu2020cost}}, optimising locations and qualities of infrastructure~\mbox{\cite{wu2021optimizing,wu2020iot,meng2017determining}}, and facility planning in healthcare ~{\cite{zuo2019optimizing}}. Particularly in the EV context, most of the existing work focuses on the deployment of charging facilities. For instance, the work in~{\cite{Li:ICDE:2015}} proposes an approach to find locations to deploy chargers using electric taxi trajectory data, while~{\cite{liu2018integrated}} considers an integrated multi-criteria decision making approach for charging stations planning. The work in~{\cite{Xiong:TITS:2017}} also considers the cost for charging when optimising charger locations, while~{\cite{Liu:SIGSPATIAL:2019}} proposes two heuristic-based algorithms to deploy EV chargers given their potential social benefits. It also considers the cases of incremental deployment, but they essentially re-run the proposed algorithms on the new set of candidate locations. Our work is different in that we have different objectives to optimize. These work considers the deployment public chargers across the city, which aims to satisfy as much charging demand as possible with a given budget on charger deployment. In our case besides service coverage, we also care about profit as a private EV sharing service provider. In addition, our approach does not need to work with a predefined budget, which is often difficult to estimate in practice, but just finds the self-sustaining deployment plans that can cover the cost. On the other hand, the work in~{\cite{Du:KDD:2018}} also studies EV charger planning for private EV sharing platforms, which shares the similar problem with our work. However, it assumes the deployment of chargers is an one-off task, and does not consider dynamic deployment cases as studied in this paper.}

\noindent \textbf{Neural Search and Optimization. }
Recently, there is an emerging interest in using deep neural networks to solve a broad range of optimization problems. Given their structure, it is often possible to formulate the optimization problems as sequential decision progresses and use a single or multiple reinforcement learning agents~\cite{bertsekas2021multiagent}, which learn the heuristics implicitly to find a solution. For instance, the work in~\cite{barrett:arxiv:2019} proposes a DQN based approach to address the Max-Cut problem, which becomes the new state-of-the-art. In~\cite{cappart:AAAI:2019} the authors design a general RL approach for combinatorial optimization and show promising results on the Maximum Independent Set Problem. The work in~\cite{huang:arxiv:2019} tackles the Graph Coloring Problem for large graphs by using efficient network architectures to speed up optimization. \rev{Besides the traditional optimization problems, neural search techniques have also been widely used in the emerging AutoML field, which aims to automate the machine learning pipeline~{\cite{zoph:ICLR:2016}}, or combining the strengths of different models via imitation learning~{\cite{kebria2019deep}}.
Our work bears close resemblance to those existing work, 
however, we try to solve a different problem in the shared e-mobility context, and propose a novel hierarchical controller architecture to search efficiently, which has not been considered by the existing work.}

\section{Simulation Design}
\label{sec:simulation}

\rev{In this section, we present our design of the high-fidelity simulation environment for shared e-mobility systems at city scale, which builds the playground for the proposed multi-agent neural search algorithm to learn, by trial-and-error, how to optimize deployment with specific goals (e.g., revenue vs. service coverage) in mind. Our simulation environment has two key advantages. Firstly, it is able to simulate the complete operation process of the shared e-mobility system, from users picking up an EV, to returning the EV to destination and charging, at a fine-grained granularity, capturing the unique properties of EV sharing such as range limitations, and the complex interactions between users and the system. Secondly, we calibrate our simulation with multi-modal data collected from the real world, ensuring that it can faithfully capture the key characteristics of the actual system, and thus exhibits similar behaviour as the system operated in practice.}



\subsection{Overview of Shared e-Mobility Systems}
\label{sub:simu-overview}

\noindent \textbf{The Shared e-Mobility Model.}
\rev{In this paper we consider \emph{station-based} e-mobility systems, i.e., the users have shared access to the EVs of the system, but can only pick up EVs from and return them to the available stations. Let $s$ be such a station. In our case, $s$ can be represented as a tuple $(\texttt{loc}, \#\texttt{c})$, where \texttt{loc} denotes the location (e.g., latitude and longitude coordinates) of $s$, and \#\texttt{c} is the total number of charging docks (equivalent to parking spaces) within the station $s$. In practice, the system may deploy new stations or close existing ones across the city. When a station $s$ is firstly deployed for operation, a number of EVs (denoted by \#\texttt{v}) would be assigned to $s$, where typically $\#\texttt{v}<\#\texttt{c}$. This means at the beginning the station $s$ should have \#\texttt{v} EVs available for pick up and \#\texttt{c} - \#\texttt{v} free spaces for potential EV returns.}

In such shared e-mobility systems, the vehicles are of limited range.
In this work we assume the fully charged ranges of the EVs for given vehicle models are fixed, and during normal driving the remaining range can be determined by their discharging curves~\cite{Tremblay:Charging:2009}. In practice, we observe that the time for charging of the EVs is much longer than refilling the traditional vehicles, which however can be estimated by the corresponding charging models~\cite{Tremblay:Charging:2009}, given the remaining range, battery capacities and charger specifications. 

\rev{Given the set of its available stations $S$, the EV sharing system operates as follows. Assume that at time $t$, a user wants to rent an EV from a certain station $s^{\text{o}}\in S$, and return to the destination station $s^{\text{d}}\in S$. If she finds there is at least one vehicle available at the picking up station $s^{\text{o}}$, that has been sufficiently charged with enough range to cover her planned trip, she will post an order $o_t = (s^{\text{o}}, s^{\text{d}})$ to the EV sharing system. Upon accepting the order, the system allows the user to take over the EV. The order is considered to be completed (i.e., the demand of this user is satisfied) when the EV is returned to the destination station $s^{\text{d}}$ and plugged in for charging, which is considered to be ready for the next user. The price for this order is calculated based on her rental duration. If the destination does not have a vacant parking space, we assume the system should reposition the EV to the nearest station which has parking/charging availability within certain radius (3km in our simulation), which is often achieved by a team of dedicated staff in real world operations.}

\begin{figure*}[t]
 \centering
 \begin{subfigure}{0.6\columnwidth}
 \includegraphics[width=\textwidth]{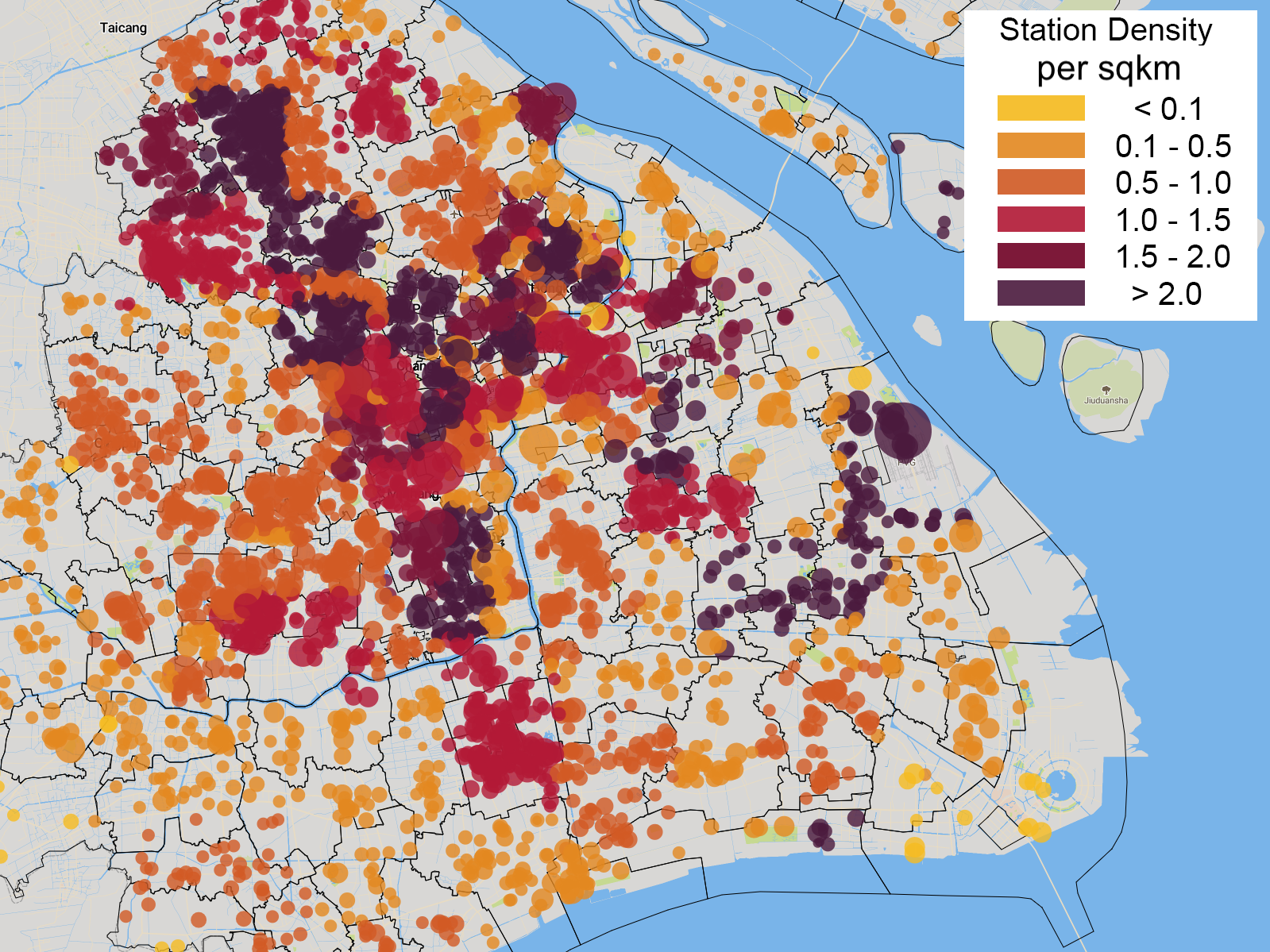}
 \caption{}
 \label{fig:spatial-stations}
 \end{subfigure} 
 \begin{subfigure}{0.6\columnwidth}
 \includegraphics[width=\textwidth]{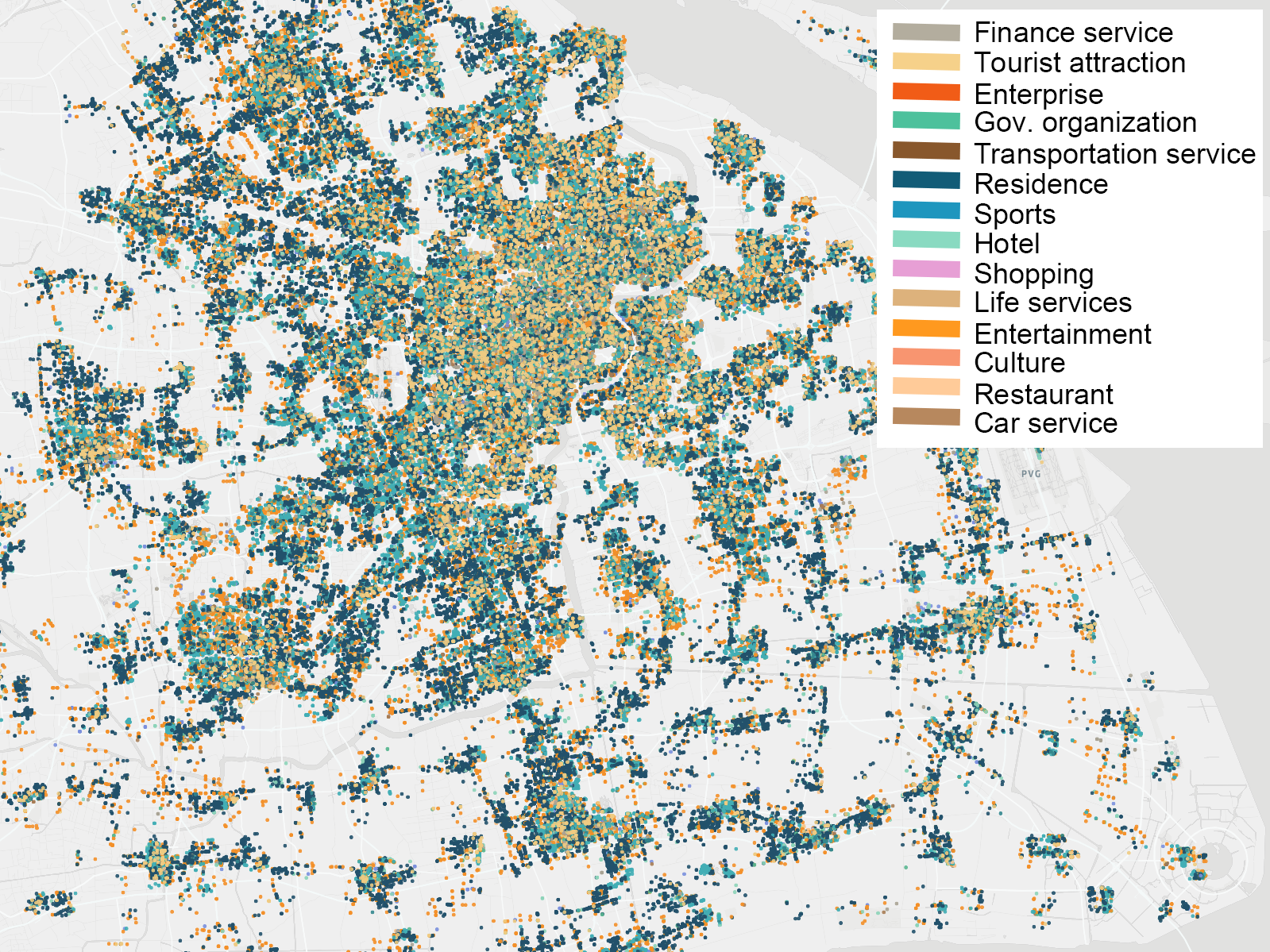}
 \caption{}
 \label{fig:spatial-poi}
 \end{subfigure} 
 \begin{subfigure}{0.67\columnwidth}
 \includegraphics[width=\textwidth]{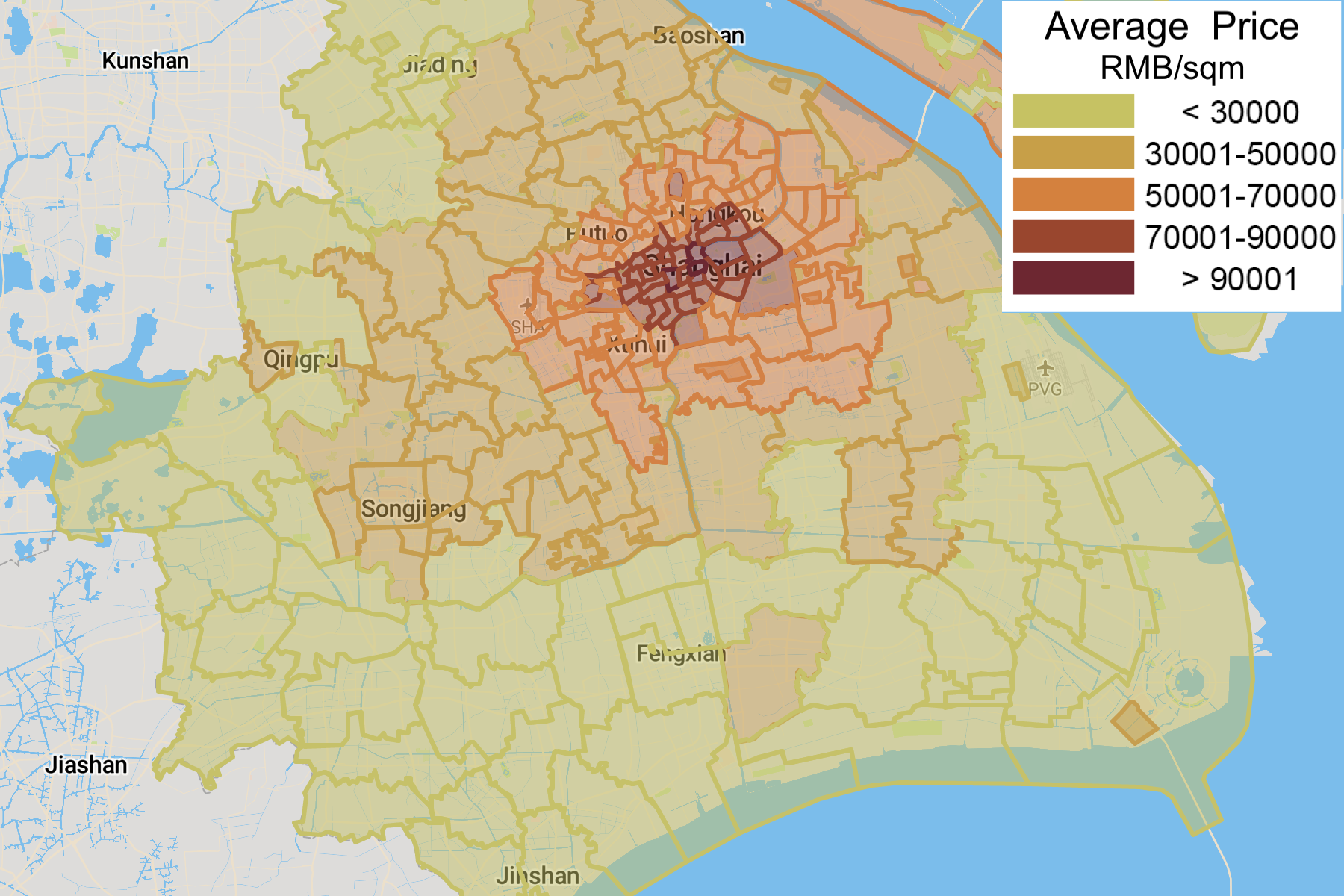}
 \caption{}
 \label{fig:spatial-price} 
 \end{subfigure} 
 
 \caption{Visualization of the collected data. (a) The distribution, density and sizes (\# of parking spaces/charging docks) of the stations in the real-world shared e-mobility system. (b) POI distribution in the city. (c) Average property price across different regions in the city. }
 \label{fig:spatial}
 \end{figure*}

\noindent \textbf{The Shared e-Mobility Data. }
\rev{We have collected a rich set of data from a real-world shared e-mobility system in Shanghai for 12 months to support and calibrate our simulation environment, ensuring its high-fidelity with respect to the systems operating in the real world. More concretely, the dataset contains a) the complete \textbf{transactions of orders} in the system, i.e., when and where a user rented/returned a vehicle, and b) information on \textbf{station deployment}, i.e., when and where a station was deployed or closed. In the order data, each record contains detailed information about this order, including the anonymous user ID who initiated the order, the ID of the origin/destination stations, the timestamps of pick-up/return, the total duration of the order and the final price etc.} In total we have collected $>$7 million valid order transactions, which were generated by approximately 0.4 million active users during the one year period. For the station deployment data, we have collected the service status at each day during the 12 months period, including station locations, numbers of charging docks in each station (see Fig.~\ref{fig:spatial-stations} for visualization), and when and where stations were deployed/closed. \rev{We also collected additional auxiliary data to support our simulation, e.g., to calculate the cost for infrastructure deployment, estimate the potential service coverage, and infer the demand for travel. This includes the POI data of the city~{\cite{Web:AMap}} (see Fig.~{\ref{fig:spatial-poi}} for visualization), the average property prices across different regions~{\cite{Web:lianjia}} (see Fig.~{\ref{fig:spatial-price}} for visualization), taxi trajectory data during the same period of time~{\cite{Web:TaxiSODA}}, and the charging/discharging models of the electric vehicles~{\cite{Tremblay:Charging:2009}}.}

\subsection{Simulating Dynamic Deployment of Stations}
\label{sub:simu-stations}
We construct our simulation environment by first simulating the deployment in the stations of the shared e-mobility system. 
For simplicity, in our simulation we assume that any stations to be deployed are drawn from a candidate pool $\mathbb{S}$, containing all possible candidates of stations in the system. Without loss of generality, we consider a 2D world where a station candidate $s \in \mathbb{S}$ is a point at location \texttt{loc} and with \#\texttt{c} charging docks. In our implementation, we generate the candidate pool $\mathbb{S}$ according to the station deployment data collected from the real-world system, i.e. for any station $s$ that has been deployed once in that system, regardless of whether it had been closed or not, we add $s$ to the pool $\mathbb{S}$. In this way we obtain the candidate pool of stations for deployment in simulation, which forms the search space for the proposed approach. 

\rev{We consider \textit{dynamic deployment} of stations, i.e., stations can be opened or shut down at arbitrary times. Without loss of generality, we assume the time granularity for station deployment is days, i.e., they can be adjusted on daily basis (practicality of this paradigm in the real world discussed later in Sec.~{\ref{sec:discussion}}).} Let $S_t \subseteq \mathbb{S}$ be a subset of the candidate pool. We refer to $S_t$ as a \textit{deployment snapshot} at time $t$, and assume by the end of the timestamp only the stations in $S_t$ will remain active in simulation. The sequence of such deployment snapshots from the beginning to the end of the simulation episode is defined as the \textit{deployment plan} $\boldsymbol{S}$, where $\boldsymbol{S} = \{S_t\}$, $t\in[1, T]$. In essence, $\boldsymbol{S}$ describes the evolving process of the stations in the shared e-mobility system, which as discussed later is the key for performance, such as the revenue and demand satisfaction rate of the service.

We also assume that for each candidate station $s$, there is a cost $c(s)$ for being active in operation of the system per unit time. In our simulation, this cost is set based on two factors: i) the property price around $s$; and ii) the size (i.e., number of parking spaces \#\texttt{c}) of $s$. This is consistent with the cases we observed from the real-world data, where the rental prices of stations depend on their sizes, e.g., those with more parking spaces tend to be more expensive, and also differ in different areas. Therefore, at given time $t$ the cost of running the service depends on the current deployment snapshot $S_t$, i.e., $c(S_t) = \sum_{s\in S_t} c(s)$. Then for a deployment plan $\boldsymbol{S}$, we define the total cost in this episode $c(\boldsymbol{S})$ as the sum of the costs incurred by its snapshots: $c(\boldsymbol{S}) = \sum_{t = 1:T} c(S_t)$. \rev{Note that here for simplicity we do not consider the upfront deployment cost such as installing chargers etc., which in practice can be factored into the long-term running cost, and assume there is no overhead of shutting down or re-opening stations. }

\subsection{Generating Spatio-temporal User Demand}
\label{sub:simu-demand}
To better simulate the operation details of the actual shared e-mobility system, we set 10min as one simulation step. Note that this is different from the timestamps considered in station deployment process as discussed above, which is a day (24 hours), containing 144 simulation steps. 
We define a particular user demand as a tuple $d=(k,\texttt{o},\texttt{d})$, where $k$ is the current simulation step, \texttt{o} and \texttt{d} are the origin and destination of this demand respectively. We represent \texttt{o} and \texttt{d} with 2D coordinates e.g., GPS locations across the space. \rev{This represents the intention of travel from \texttt{o} to \texttt{d} of the users, and in our simulation we learn to generate user demand from the real-world data. In practice, the user demand may be affected by the deployment of infrastructure itself, e.g., users may want to try the system when they find there are stations nearby. To mitigate this, in our simulation we consider two independent data modalities, i) the historical order transactions of the shared e-mobility system, as well as ii) the taxi trajectory data (cleaned to only have origins and destinations). We use Gaussian Processes (GP) with RBF kernels to fuse the data from both modalities, approximating the genuine ``intend to travel'' of users at certain point and time. In particular, we first consider the pick up demand (i.e., the origins \texttt{o}) and learn GP models over space and time, with which at each simulation step $k$ we can obtain a spatial distribution of the potential origins in user demand. This also means for any point in the 2D space, we have an estimate of the numbers of users that intend to travel from that point at step $k$, with certain confidence interval.}

When generating the demand in simulation, at given step $k$ we sample the learned GP over the candidate station pool $\mathbb{S}$, i.e., at each candidate station we obtain the amount of travel demand starting from that location at step $k$. Here we only sample over $\mathbb{S}$ rather than the entire 2D space because eventually user demand will be satisfied at those stations, while the GP model has already taken the neighbourhood knowledge into account, e.g., nearby taxi demand or EV pick-up demand at stations within certain distance. Therefore in the following text, we consider the user demand only at the candidate stations, i.e., $d=(k,s^o,s^d)$, where $s^o,s^d\in \mathbb{S}$. 

\rev{To generate the complete demand in pairs of origins and destinations, for a user demand originated from $s^o\in\mathbb{S}$, we use another GP to approximate the spatio-temporal distribution of possible destinations from the historical orders of the e-mobility system as well as taxi data. In essence, at step $k$ for this demand with origin $s^o$, the GP model generates a 2D distribution over the space, and by sampling from the distribution over the candidate pool $\mathbb{S}$ we can pin down the destination station $s^d$ for this particular demand. The reason why we adopt this two-phase approach rather than sampling independently from the distributions of origins and destinations, is that in practice the origins and destinations of user demand are highly correlated, and the correlations often depend on different times. In addition, unlike existing work~{\cite{Du:KDD:2018}} which directly uses historical data as demand, we use GPs to i) fuse different sources of information (taxi and the e-mobility data), and ii) generate distributions of user demand rather than scalar values for robustness. As shown later in Sec.~{\ref{sec:eval}}, this improves the fidelity of our simulation, and helps the proposed algorithm to generalize better.}

\subsection{Operating Shared e-Mobility Systems in Simulation}
\label{sub:simu-operation}

For each simulation episode, let $\boldsymbol{S}$ be a deployment plan of EV sharing stations, e.g., generated by our search algorithm. In essence, $\boldsymbol{S}$ describes which candidate/existing stations should be deployed/removed from the service and when. We assume that this station deployment/adjustment process happens at the beginning of a timestamp $t$ (0am on each day), which updates the active stations according to the corresponding snapshot $S_t$ in the deployment plan $\boldsymbol{S}$. Note that any candidate station to be deployed, we allocate certain number of EVs (fully charged) to this new station for its future operation, according to its maximum capacity (\#\texttt{c}) and a probability ratio learned from the historical data. Typically in our simulation we assign 0.5\#\texttt{c} to 0.7\#\texttt{c} of EVs to the new stations, leaving sufficient parking spaces for incoming EVs from other stations.

\rev{Then for each simulation step $k$ (there are 144 steps per time $t$), our simulator generates user demand as discussed in the previous section. It then operates the system in the same way as discussed in Sec.~{\ref{sub:simu-overview}}, i.e., processing the simulated user orders, handling vehicle charging/returns. Then at time $t$, with the deployment snapshot $S_t$, we define the total revenue value at $t$ as $\texttt{GMV}(S_t)$, which is calculated over all the satisfied orders of the EV sharing service during that time. The net revenue value $\texttt{NV}(S_t)$ is then defined as GMV subtracts the deployment cost: $\texttt{NV}(S_t) = \texttt{GMV}(S_t)-c(S_t)$. For deployment plan $\boldsymbol{S}$, we define its GMV and net revenue as their summations over $T$: $ \texttt{GMV}(\boldsymbol{S}) = \sum_{t=1:T}\texttt{GMV}(S_t)$, and $ \texttt{NV}(\boldsymbol{S}) = \sum_{t=1:T}\texttt{NV}(S_t)$ respectively. }

\section{Problem Formulation}
\label{sec:method}





\subsection{The Deployment Optimization Problem}
\label{sub:problem}
Let us assume at time $t=0$, the simulation world is at the initialization state, i.e., the shared e-mobility system has some stations deployed and is ready to operate. This is reasonable since in practice, such systems tend to deploy their first batch of stations before going live to the public. Let $T$ be the length of one simulation episode, i.e., we run the simulator for $T$ timestamps (days in our case) in one training/testing pass. As discussed above, we assume during the $T$ days the stations of the EV service can be dynamically deployed or removed on a daily basis, i.e., at each $t$, according to a deployment plan $\boldsymbol{S} = \{S_t\}$, $t\in[1, T]$, where $S_t$ is the set of stations to be active at $t$, drawing from the candidate pool $\mathbb{S}$.

Let $\mathcal{P}(\boldsymbol{S})$ be the set of all possible deployment plan of length $T$ for this simulated e-mobility system. Then our deployment optimization problem is to find the optimal deployment plan $\boldsymbol{S}^*\in \mathcal{P}(\boldsymbol{S})$:

\begin{equation}
    \label{eq:obj}
    \begin{aligned}
        &\boldsymbol{S}^* = \; \underset{\boldsymbol{S}\in \mathcal{P}(\boldsymbol{S})}{\argmax} \, \texttt{SC} (\boldsymbol{S}) + w\texttt{PM}(\boldsymbol{S})\\
        &\textrm{s.t: }   \quad c(S_t) \leq B(t), \, \text{where } S_t\in \boldsymbol{S}, \forall t \in [1:T].
    \end{aligned}
\end{equation}

\noindent Here $\texttt{PM}(\boldsymbol{S})$ is the \textit{profit margin ratio} of the deployment plan $\boldsymbol{S}$, i.e., the percentage of profit (net revenue \texttt{NV}) with respect to the total income (\texttt{GMV}) if deploying stations as instructed by $\boldsymbol{S}$. $\texttt{SC}(\boldsymbol{S})$ is the \textit{service coverage ratio} of the deployment plan $\boldsymbol{S}$, which is defined as the normalized sum of two ratios: i) the percentage of satisfied user demand, and ii) the percentage of POIs covered by the service. We assume a POI is covered by a station if it is within 1$km$ radius of the station. Therefore, at time $t$ the service coverage of the corresponding deployment snapshot $\texttt{SC}(S_t)$ depends on both satisfied demand and covered POIs. Similar to $\texttt{NV}(\boldsymbol{S})$ defined above, we define $\texttt{SC}(\boldsymbol{S})$ as the mean service coverage of this particular deployment plan over $T$ timestamps, $\texttt{SC}(\boldsymbol{S}) = T^{-1}\sum_{t=1:T}\texttt{SC}(S_t)$. In our case, both objectives are in percentage, and thus we can use a weight $w\in[0,1]$ to balance them, which can be tuned for scenarios with different priorities, e.g., aiming to roll out service to more users vs. obtain more profit. \rev{The constraint requests that the instantaneous cost of the deployment plan at $t$ should not exceed the value of a budget function $B(t)$, which can be fixed or varying over time. Unlike existing solutions which typically assume fixed budget (a total sum or evenly spread), in this work we consider a more realistic budget function, $B(t) = \texttt{GMV}(S_t)$, which requires the system to be \textit{self-sustaining}. In particular, this means that under a given deployment plan $S_t$, the income $\texttt{GMV}(S_t)$ of the system at any $t$ should be able to cover the cost of infrastructure deployment, i.e., we want the system to be break-even. }

In fact, if we unfold the deployment plans over time, this deployment optimization problem is essentially a constrained combinatorial optimization problem. However, the search space in our case can be prohibitively large, e.g., with 4$k$ candidate stations over 30 days period there will be $2^{4k^{30}}$ possible plans to evaluate. In addition as discussed in previous sections, the impact of deploying/removing a particular station is often complicated, i.e., we won't be able to directly estimate the gain of such an action (e.g., increase in \texttt{NV} and \texttt{SC}) without running the system through. This makes most of the existing heuristics-based optimization approaches~\cite{Du:KDD:2018,Liu:SIGSPATIAL:2019} infeasible, as they assume the benefits/utility (e.g., of deploying a particular charger/station) are independent and known a priori. In the next section, we show how we formulate this problem as a Multi-agent Reinforcement Learning (MARL) task, which allows us to explore the search space via trial-and-error.

\subsection{Deployment Optimization as a MARL Task}
\label{sub:marl}
In this paper, we consider a set of autonomous agents, each of which interacts with the common environment to improve its behavior. Comparing to the single agent settings, the multi-agent formulation is more suitable to address our problem, because multiple agents could better exploit the decentralized nature of the station deployment task. By design learning can be more efficient than the single agent case, as the action spaces of the agents are much smaller, and the computation can be accelerated by parallel processing. Given the environment (i.e., EV sharing service), we model the deployment optimization problem as a Markov Game $G = (N, \mathcal{X}, \mathcal{A}, \mathcal{T}, \mathcal{R}, \gamma)$, in which at most $N$ agents interact with the environment, making decisions as to where and when stations should be deployed. $\mathcal{X}$ represents the states of $G$ and $\mathcal{T}$ is the transition function between states. $\mathcal{A}$ is the joint actions of the agents, $\mathcal{R}$ is the reward function, and $\gamma$ is the discount factor. 

\noindent \textbf{Agents. }
In our multi-agent formulation, we assume an agent controls the deployment of EV stations within a certain geospatial region. Essentially, we delegate the agent to manage the station deployment process of that region, who decides which candidate stations should be deployed or closed and when. We partition the space into regions belong to different agents by clustering candidate stations locations in $\mathbb{S}$. In particular, we cluster the candidate stations based on their pairwise distances, where boundaries of regions are formed by convex hulls of clusters. Let us assume we have obtained $N$ regions as the result of clustering, managed by $N$ agents.

\rev{To make sure that the agents face similar learning load, we also require each cluster to have the same amount of $M$ candidate stations. In practice, this may lead to regions with different sizes due to the spatial variation in the density of candidate station locations, e.g., we observe that the obtained regions in city center is much smaller than those at the city edges. However, we found that this won't affect the agents' performance and behaviour, since the candidate station density is also highly correlated with POI density and distribution of potential user demand. This means although managing regions with different sizes, the agents tend to learn to cover similar levels of demand/POIs with its deployment plans, only at different spatial scales. It is also worth pointing out that our approach can also work with the other space partition paradigms, such as hexagonal or rectangular grids, as long as each region/grid is managed by a individual agent.}

\noindent \textbf{States and Observations. }
At time $t$, for the $i$-th region, its state $x_t^i$ encodes information about the user demand, EV status and station deployment within the boundary of this region. In particular, we consider both the set of currently online stations, and candidate stations that are not yet deployed. For each of them, we include its location \texttt{loc}, number of available charging docks \#\texttt{c}, the deployment cost, the numbers of EVs parked in the stations and their individual range at each simulation step within $t$, as well as the potential future rent/return requests (number of EVs) and the average value of potential future orders in the next timestamp. The global state $\boldsymbol{x}_t$ is the combination of states of each region $\boldsymbol{x}_t = \{x_t^i\}$, $i\in[1, N]$. \rev{Let agent $i$ manage the $i$-th region. At time $t$, we assume the agent observes both its local state $\{x_t^i\}$ and the global state $\boldsymbol{x}_t$. This allows the agents to observe and interact with not just its local environment, but also encourages them to learn to better cooperate with the others that have connections with them, e.g., agents of neighbouring regions, or those with high correlations in terms of user demand (i.e., users may frequently travel from one region to another).}

\noindent \textbf{Agent Actions and State Transitions. }
For agent $i$, its action $a_t^i$ describes the deployment snapshot of the $i$-th region, i.e., which stations should be deployed or removed. Therefore at time $t$, a joint action $\boldsymbol{a}_t \in \mathcal{A}_t^1 \times ... \times \mathcal{A}_t^{N_t}$ all the $N$ agents forms the complete deployment snapshot $S_t$, as introduced above. Upon performing the joint action at $t$, the current state $\boldsymbol{x}_{t}$ will transit to the next state $\boldsymbol{x}_{t+1}$ according to the state transition probabilities $\mathcal{T}$, which are defined as $\mathcal{T}(\boldsymbol{x}_{t+1}|\boldsymbol{x}_{t}, \boldsymbol{a}_t)$. Note that in many MARL cases, it is often not possible to describe the state transitions as functions with analytical forms, and thus in our case we do not attempt to model $\mathcal{T}$ explicitly, but rely on our simulator to capture the state transitions. 

\noindent \textbf{Reward Function. }
For each agent, the reward $r_t^i$ of taking an action at time $t$ is determined by the reward function: 
\begin{equation}
    \label{eq:general-r}
    \mathcal{R}^i(\boldsymbol{x}_t, \boldsymbol{a}_t): \mathcal{X}\times \mathcal{A}_t^1\times ...\times \mathcal{A}_t^{N_t} \rightarrow \mathbb{R}
\end{equation}
As discussed above, in our deployment optimization problem, we would like to maximize both the profit margin (\texttt{PM}) and the service coverage (\texttt{SC}). In practice, we find that the two objectives often diverge, e.g., maximizing \texttt{PM} would often lead to greedy agents that always try to deploy new stations at ``hot'' locations with high profit. On the other hand, optimizing \texttt{SC} tends to encourage the agents to spread stations across the space, while keeping many less profitable stations active, resulting in decrease in net revenue. In addition, we also require the service to be sustainable, in that the GMV of the deployed stations should be able to cover their cost. To balance such factors, given the agent action $a_t^i$, which requires the set of stations $S_t^i$ to be active at time $t$, we design the reward function $r_t^i$ based on our optimization objectives as in Eq.~\eqref{eq:obj}:
\begin{equation}
    \label{eq:agent-r}
    r_t^i = g^{\texttt{\,SC}}\big(S_t^i,S_{t-1}^i\big) + w \, g^{\texttt{\,PM}}\big{(}S_t^i,S_{t-1}^i\big{)} + \lambda \, \text{min} \{\texttt{NV}(S_t^i), 0\}
\end{equation}
where $g^{\texttt{SC}}(S_t^i,S_{t-1}^i)$ and $g^{\texttt{PM}}(S_t^i,S_{t-1}^i)$ calculates the \textit{improvement rate} in terms of service coverage \texttt{SC} and profit margin \texttt{PM}, given the current deployment snapshot $S_t^i$ and the previous one $S_{t-1}^i$. $w$ is the weight balancing the two objectives, as in Eq.~\eqref{eq:obj}. The term $\lambda \, \text{min} \{\texttt{NV}(S_t^i), 0\}$ penalizes any $S_t^i$ that produces negative net revenue with a scaling factor $\lambda$, which is learned via grid search. Given the reward function, each agent aims to maximize its discounted reward $\mathbb{E} [ \sum_{k=0}^{\infty} \gamma^{k} r_{t+k}^i ]$, where $\gamma \in [0,1]$ is the discount factor. 

\begin{figure*}[ht!]
\centering
\includegraphics[width=0.7\textwidth]{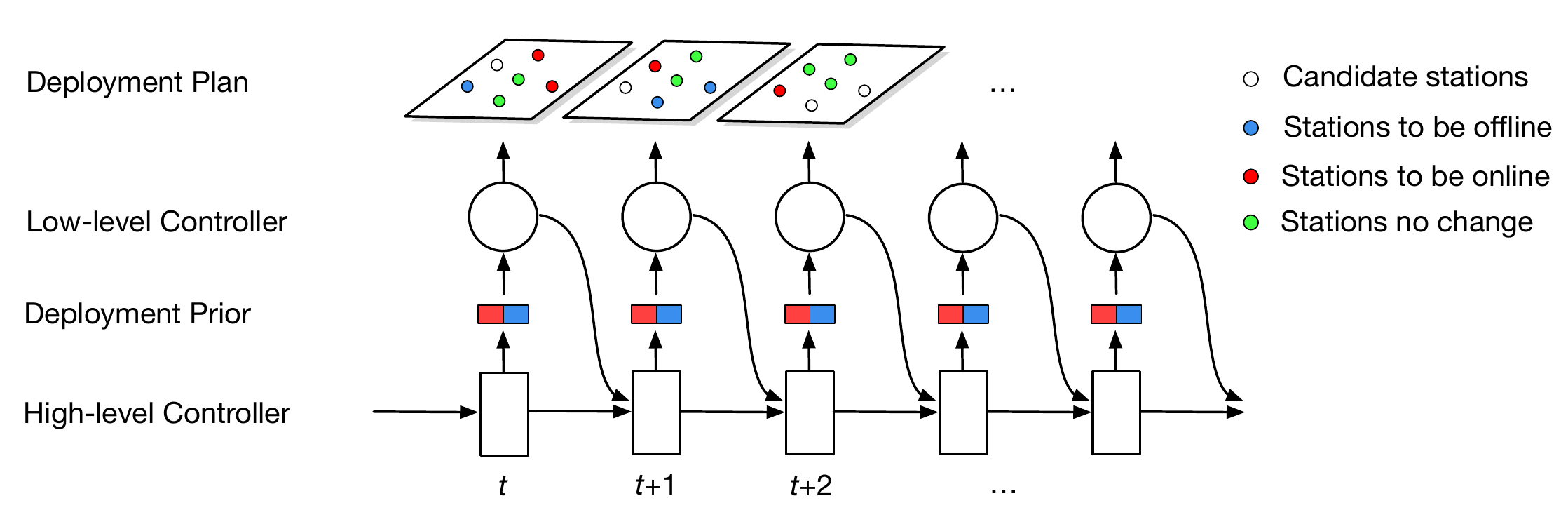}
\caption{The architecture of the proposed hierarchical controller.}
\label{fig:controller}
\end{figure*}

\section{Hierarchical Deep Neural Search}
\label{sec:search}
Under the above MARL formulation, we design a novel hierarchical neural search algorithm, which learns to generate better deployment plans over time. The key idea is to use a controller network to generate various deployment plans (i.e., actions), and evaluate the performance of those plans in our simulation environment. The reward/penalty signals collected from the simulation are then propagate back and used to update the controller with reinforcement learning paradigm, so that it tends to produce improved plans in future runs. 

\subsection{Hierarchical Controllers Design}
\label{sub:hierarchical}
In this paper, we consider a LSTM-based controller, which recursively generates deployment plan $S_t$ at each $t$ for $T$ timestamps. In fact, given the multi-agent setting, for each agent $i$, we can use a boolean string of length $M$ to represent $S_t^i$, where $M$ is the total number of candidate stations in the regions. Such encoding has been widely considered in tasks such as searching for neural network architectures. However unlike the existing work, in our case even with multi-agent formulation, directly generating and searching from the $2^M$ sized encoding space is prohibitively expensive. To tackle this challenges, we design a hierarchical controller which generate $S_t^i$ progressively, as shown in Fig.~\ref{fig:controller}. The key idea is that instead of directly sampling from the distribution of deployment plans at each $t$, we generate $S_t^i$ in two stages. Concretely, we design a hierarchical structure with two connected controllers: a \textit{high-level} and a \textit{low-level} controllers respectively. 

\noindent \textbf{High-level Controller.}
Given the previous and current states, rather than delving into deployment details, the high-level controller only generates a categorical \textit{deployment prior} for the region $i$ at time $t$. In particular, we consider a deployment prior as a pair ($S_t^{i+}$, $S_t^{i-}$), where $S_t^{i+}$ is a categorical variable indicating the amount of new stations to be made online during time $t$ at region $i$, e.g., 10\% of the total candidates in the region, while $S_t^{i+}$ is for the stations to be made offline. In this case, if the domain of the variables is limited, then for the high-level controller the search space at each timestamp $t$ is manageable. On the other hand, such a deployment prior indicates the coarse volume of the desired deployment plan for the region at time $t$, e.g., the high-level controller may prefer to deploy more rather than removing in some cases since it predicts demand would surge.

\noindent \textbf{Low-level Controller.}
Now we explain the design of the low-level controller, which translates the received deployment prior into the concrete deployment plan $S_t^i$. Essentially, now the low-level controller needs to find two sets of stations, one to deploy and one to close, while the cardinalities of them are known. In this case, one straightforward way is to continue to sample the constrained search space which is already much smaller, and generate the deployment plan $S_t^i$ accordingly. However, we found that in practice this would still lead to many meaningless trials without converging. Therefore, in this work we adopt an efficient approach to generate $S_t^i$. Concretely, for each candidate station $s_j^i$ within region $i$, we compute a score which is the combination of two values: i) its service coverage, i.e., the ratio that it could contribute to the total service coverage in this region, and ii) the ratio of expected demand it could satisfied with respect to the total demand in region $i$. Here we use a Graph Convolutional Neural Network (GCN) (e.g., as in~\cite{Luo:IMWUT:2020}) to predict the expected demand for $s_j^i$ at $t$, given the previous states in this simulation episode. The GCN is trained with the historical order data, and is able to capture the dynamics caused by station deployment/closure. 

Now we obtain the ranked list of the candidate stations in region $i$ based on their scores. Intuitively, we should deploy the best station candidates if they are not yet in operation, and close the poorly performing ones. To generate such $S_t^i$ at $t$, our low-level controller adopts a $\epsilon$-greedy approach for both deploying and closing stations. In particular, it iteratively chooses the current best/worst station candidates with the probability $1-\epsilon$, while making random selections for the rest, until the required number is reached. In our experiments, we set $\epsilon$ to 0.1. In this way, the low-level controller balances exploration and exploitation in searching of $S_t^i$, which makes our algorithm more robust to noises and perturbations.

\subsection{Training with Reinforcement Learning}
\label{sub:training}
\rev{In this paper, we consider a shared-weights controller for our agents, which is trained with reinforcement learning. As discussed in Sec.{~\ref{sub:marl}}, the generated deployment plan $\boldsymbol{S} = \{S_t\}$ is a sequence of actions $\{\boldsymbol{a}_t\}$ ($t=1:T$) to adjust the EV sharing service. By applying $\boldsymbol{S}$ in the simulation, we could evaluate the performance of $\boldsymbol{S}$, and compute the reward as Eq.~{\eqref{eq:agent-r}} and Eq.~{\eqref{eq:general-r}}. The reward signal is then fed back to the controller, to help updating its parameters $\theta$. Therefore the updated controller would have high probabilities to propose better deployment plans. Note that here $\theta$ is only relevant to the high-level controller, as the parameters of the low-level controller (the GCN) have already been trained. In our case, the reward structure is not differentiable ($\min$ in Eq.~{\eqref{eq:agent-r}}), and thus we consider policy gradient approaches to iteratively improve $\theta$. In our case, we train the controller to maximize the performance of its sampled deployment plans $\boldsymbol{S}$. The training objective can be formulated as follows:}
\begin{equation}
    \label{eq:j-func}
    J(\theta) = \mathbb{E}_{p(\boldsymbol{S};\theta)} [\mathcal{R}(\boldsymbol{S})]
\end{equation}
\rev{where $\mathcal{R}$ is the reward collected from simulation, and the expectation is taken over $p(\boldsymbol{S};\theta)$, i.e., the distribution of deployment plans $\boldsymbol{S}$ given $\theta$. In this paper we use the Proximal Policy Optimization (PPO) approach~{\cite{Schulman:PPO:2017}} to optimize $J(\theta)$, which is more sample efficient than the standard REINFORCE~{\cite{Williams:1992:REINFORCE}}, and offers faster convergence of the controller. We then iteratively optimize the objective function, until the optimal parameters $\theta^{*}$ of the controller can be found.}

\subsection{Accelerating Training with Multi-simulation}
\label{sub:multi-simulation}
\rev{As discussed above, during training each update to the controller parameters requires one run of the simulation to collect rewards, i.e., virtually operating the EV sharing system for a certain period of time. This can be time-consuming and expensive to run in practice. To speed up the learning process of the controller, we propose a multi-simulation approach that exploits the benefits of parallelism. Concretely, we set up a job queue, where the controller generates and dispatches deployment plans to this global queue. On the other hand, we employ multiple workers (across multiple threads/processes, or distributed over networked servers), each of which manages its own copy of the simulation environment. During training, a free worker picks up a proposed deployment plan from the job queue, and run the simulation to evaluate the performance of that particular deployment plan. Then the results of different workers are returned to the controller, whose parameters get updated with PPO. The controller then generates another batch of samples and this process continues until a pre-defined number of deployment plans have been evaluated.}


\section{Evaluation}
\label{sec:eval}

\begin{table*}[ht!]
\caption{Overall Performance of deployment optimization by different approaches. $\Delta$\texttt{SC} and $\Delta$\texttt{NV} are obtained with respect to Human Deployment (\textbf{HD}). For our MANS, $w$=1/9 prefers rewards in \texttt{SC}, while $w$=9/1 reward more on \texttt{NV}. Best performance values are shown in red, and second best in blue.}
\centering
\begin{tabular}{|c||c|c|c|c|c|c||@{}c@{}|@{}c@{}|@{}c@{}|} 
\hline
     & \textbf{FD}     & \textbf{HD}     & \textbf{REV}     & \textbf{COV}     & \textbf{OO}      & \textbf{IO}      & \textbf{MANS} ($w$=1)    & \textbf{MANS} ($w$=1/9) & \textbf{MANS} ($w$=9/1)  \\ 
\hline\hline
 \texttt{SC}   & 0.5511 & 0.5529 & 0.5680  & 0.6113  & 0.6054  & 0.6063  & \textcolor{red}{0.7194}  & \textcolor{blue}{0.7192}   & 0.6942    \\ 
 \hline
$\Delta$\texttt{SC} & --     & --     & 3.07\% & 10.92\% & 9.85\% & 10.02\% & \textcolor{red}{30.53\%} & \textcolor{blue}{30.50\%}  & 25.97\%   \\ 
\hline
$\Delta$\texttt{NV} & --     & --     & 11.49\% & 2.49\% & 9.86\% & 10.60\%  & \textcolor{blue}{30.92\%} & 29.82\%  & \textcolor{red}{33.55}\%   \\
\hline
\end{tabular}
\label{tbl:overall}
\end{table*}

\begin{figure*}[t]
 \centering
 \begin{subfigure}{0.5\columnwidth}
 \includegraphics[width=\textwidth]{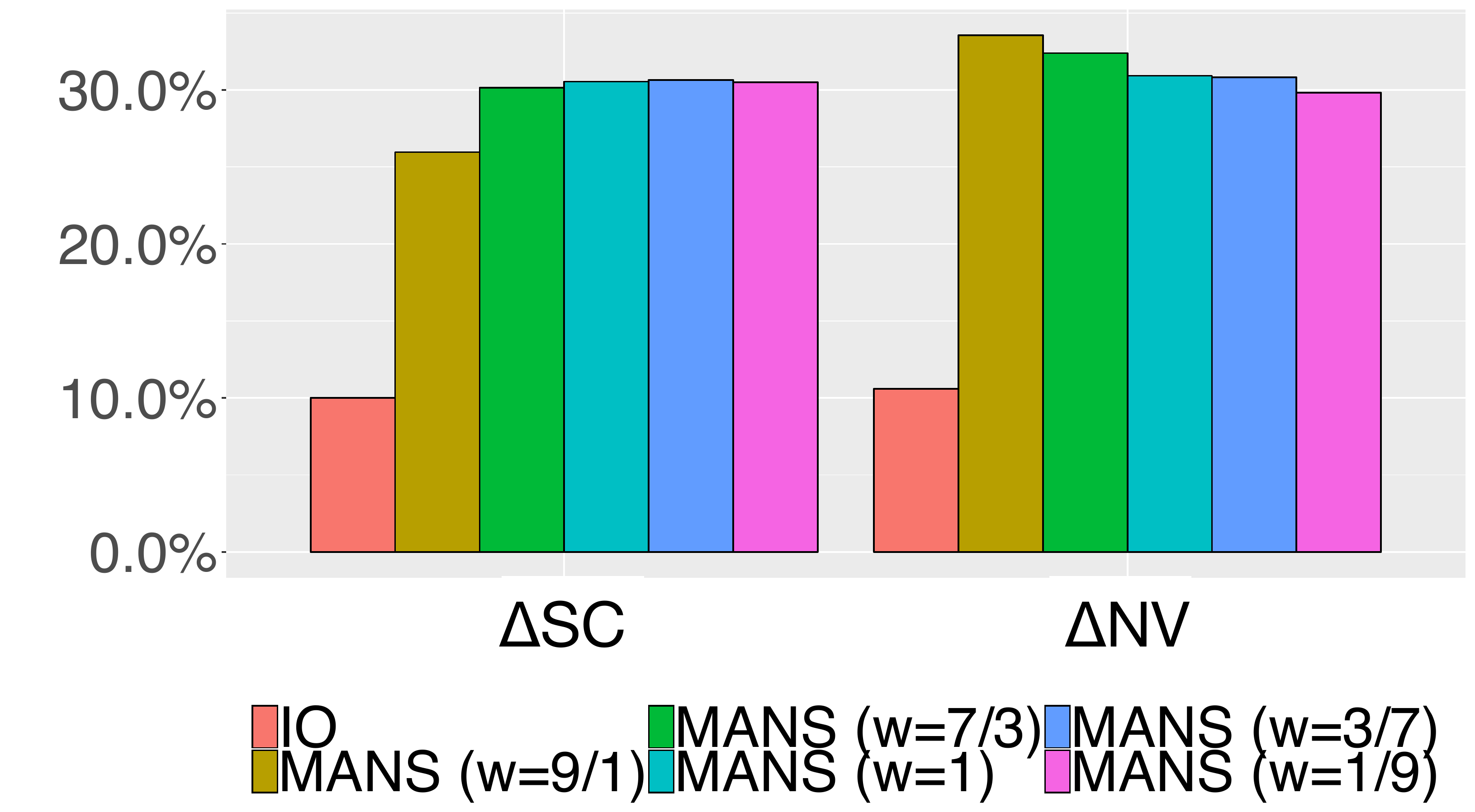}
  \caption{Impact of Weight $w$}
 \label{fig:weights}
 \end{subfigure} 
  \begin{subfigure}{0.5\columnwidth}
\includegraphics[width=\textwidth]{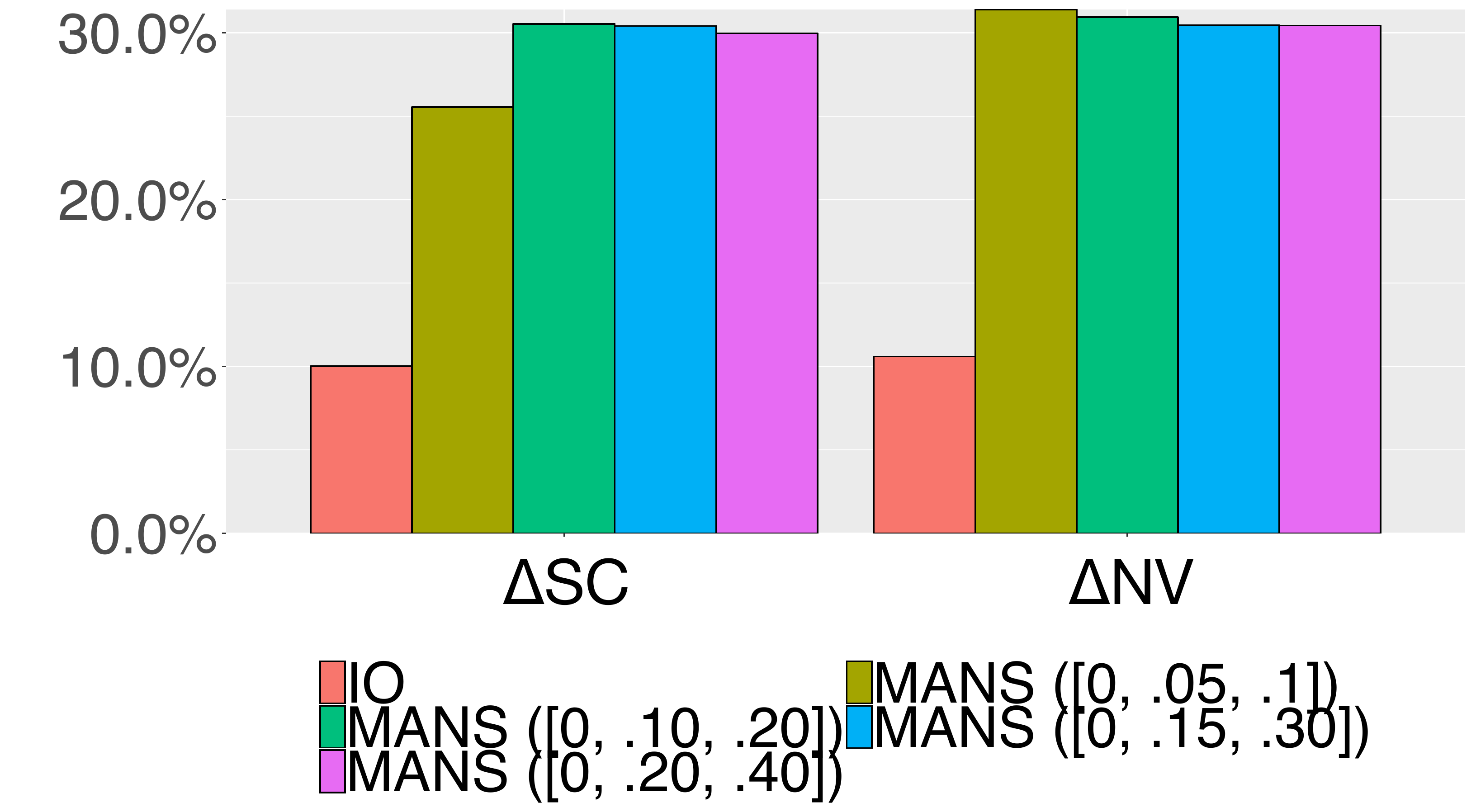}
 \caption{\texttt{SC}/\texttt{NV} vs. Action Scales}
 \label{fig:actions}
 \end{subfigure} 
 \begin{subfigure}{0.5\columnwidth}
 \includegraphics[width=\textwidth]{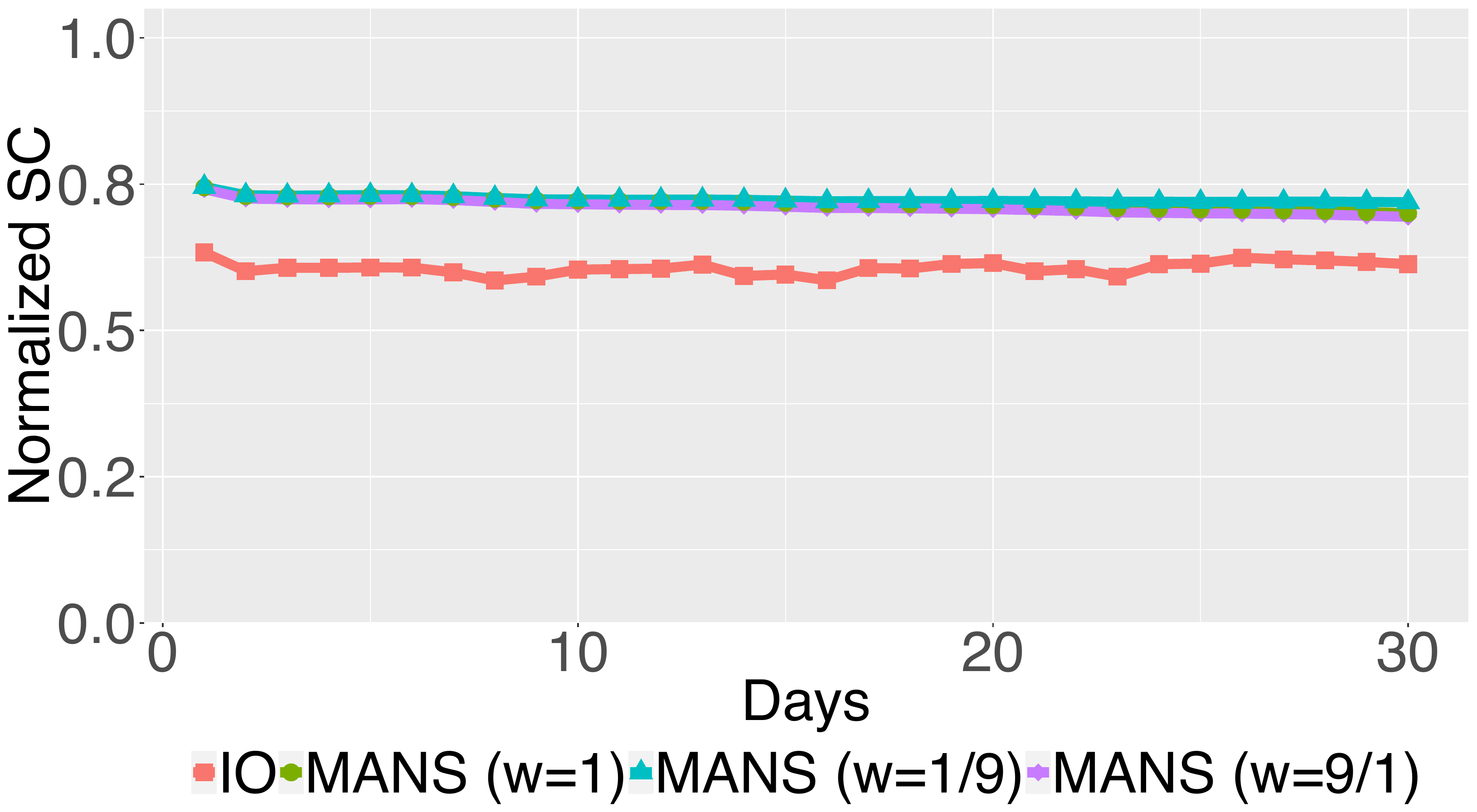}
 \caption{\texttt{SC} over 30 days w.r.t $w$}
 \label{fig:30day-sc} 
 \end{subfigure}  
 \begin{subfigure}{0.5\columnwidth}
 \includegraphics[width=\textwidth]{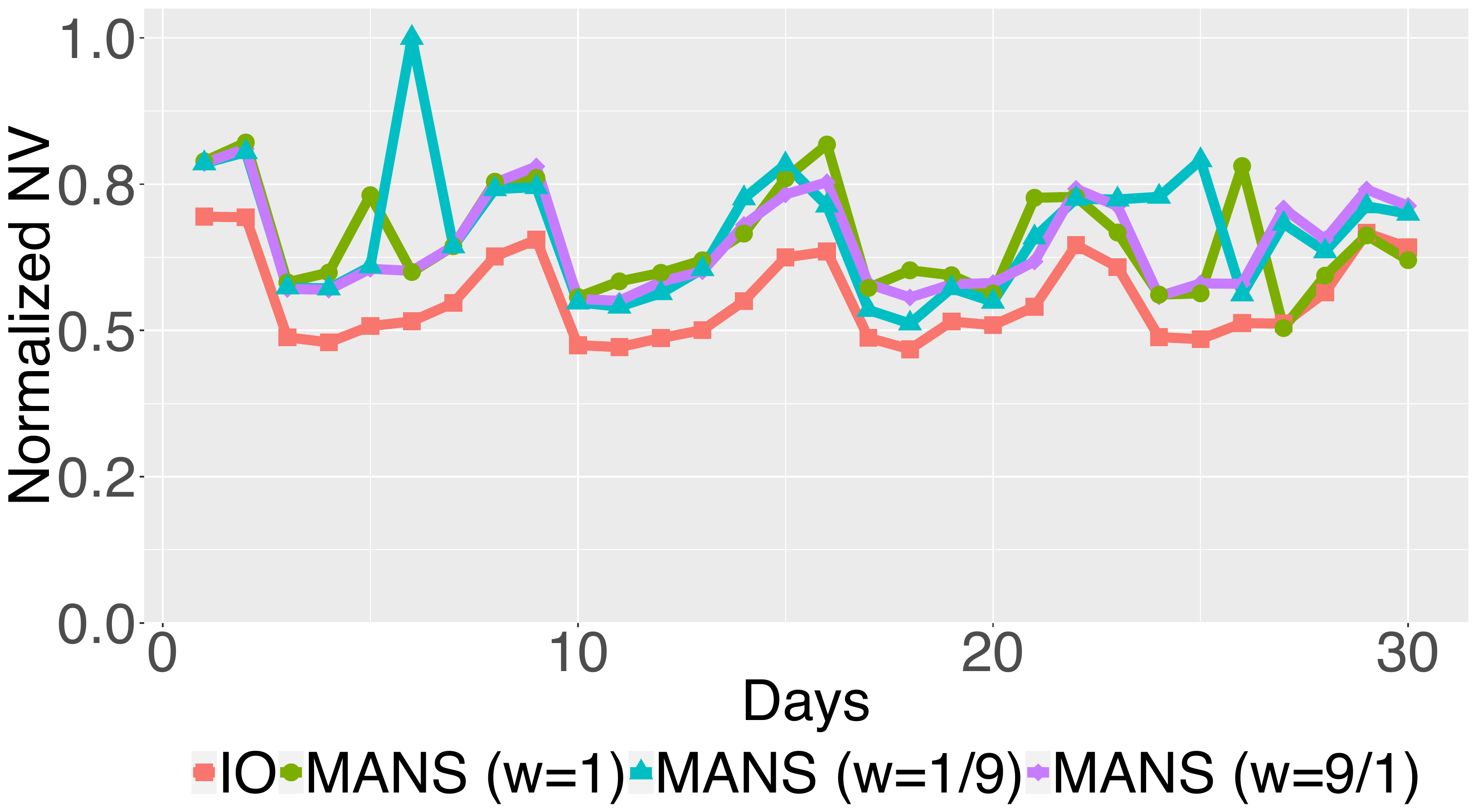}
 \caption{\texttt{NV} over 30 days w.r.t $w$}
 \label{fig:30day-nv}
 \end{subfigure} 
 \caption{\rev{Performance of the proposed MANS approach under different parameter settings (weight $w$ and action scale shown in (a) and (b)), and temporal demand distributions (shown in (c) and (d)).  Results of \textbf{IO} algorithm included as baseline.}}
 \label{fig:patterns}
 \end{figure*}


\subsection{Experimental Setup}
\label{sub:exp-setup}

We compare the proposed multi-agent neural search (\textbf{MANS}) approach with the following baseline approaches:

    \noindent \textbf{Fixed Deployment (FD)}, which keeps the stations at initialization unchanged, and there is no stations deployed/removed during the simulation.

    \noindent \textbf{Human Deployment (HD)}, which uses the deployment plan obtained from the real shared e-mobility data. Particularly, we use an independent validation dataset which hasn't been considered in simulator calibration, and take the actual deployment plan which had been implemented in the real-world. 

    \noindent \textbf{Revenue-greedy Deployment (REV)}, which selects the top/bottom stations to deploy/remove based on their historical averaged net revenue. The amount of stations to be deployed/removed is obtained by sampling a distribution learned from the validation set as in HD, i.e., the scale of changes in stations is similar with HD, but the decisions on deployment are revenue-greedy.
    
    \noindent \textbf{Coverage-greedy Deployment (COV)}, which is also heuristic-based like REV, but deploys/removes the stations according to their service coverage, i.e., it prefers to deploy stations that tend to satisfy more demand and cover more POIs. 
    
    \noindent \textbf{One-time Optimization (OO)}, which is our implementation of the the state-of-the-art charger location optimization algorithm in~\cite{Du:KDD:2018}. It assumes that the average user demand at each station can be estimated from historical data, and tries to find an one-time optimal station deployment snapshot that maximize both POI coverage and the demand satisfied rate. 
    
    \noindent \textbf{Incremental Optimization (IO)}, which is the incremental version of OO (similar to the work in~\cite{Liu:SIGSPATIAL:2019}). Unlike OO which only optimizes station deployment once and uses the solution throughout the episode, this IO performs optimization and finds the best set of stations to maintain at each timestamp. 

All the competing approaches are implemented with TensorFlow 1.14.0, and trained with NVIDIA 2080Ti GPUs. We set the same initialization of deployment snapshot $S_0$ to all the competing approaches, i.e., they start from the same situation, and try to learn the optimal deployment plan from there. As discussed in Sec.~\ref{sub:training}, we consider shared-weights controller for our agents, and train in parallel using the multi-simulation technique to improve efficiency as in Sec.~\ref{sub:multi-simulation}. We evaluate the competing approaches against the following metrics: 
    
    \noindent \textbf{Service Coverage (\texttt{SC})} as defined in Sec.~\ref{sub:problem}, which is the combination of satisfied demand rate and POI coverage; and 
    
    \noindent \textbf{Net Revenue Value (\texttt{NV})}, which is calculated as the GMV generated by the deployment plan subtracts the cost on deployment, as defined in Sec.~\ref{sub:simu-operation}. 

To be fair, instead of directly comparing \texttt{SC} and \texttt{NV} of the competing approaches, we report the improved percentages of \texttt{SC} and \texttt{NV} with respect to the baselines \textbf{HD}, indicating the performance gap between the searched plans and the actual deployment process conducted by human. 


\subsection{Experimental Results}
\label{sub:exp-results}

\noindent \textbf{Overall Deployment Optimization Performance. }
This set of experiments study the overall performance of the competing approaches. The results are shown in Table.~\ref{tbl:overall}. Firstly, we see that there is very little improvement in both \texttt{SC} between \textbf{HD} and \textbf{FD}, which means that the actual deployment plan does not offer much benefit. This is as expected because the human knowledge is only applicable to that particular case, but can't generalize to unseen scenarios in simulation. We also see the two greedy-based approaches \textbf{REV} and \textbf{COV} manage to improve \texttt{NV} and \texttt{SC} respectively, but for the other metrics which are not explicitly optimized, the improvements are marginal. On the other hand, the one-time optimization approach \textbf{OO} strikes a better balance, offering about 10\% improvement in both \texttt{NV} and \texttt{SC}. This is because \textbf{OO} uses a demand-aware heuristics, which jointly optimizes the POI coverage and the amount of demand satisfied, leading to better \texttt{SC} while also indirectly improving \texttt{NV}. The incremental optimization approach \textbf{IO} further improves the performance, in that it is able to dynamically optimize deployment plan over time, but the gap with \textbf{OO} is not significant. This is because that essentially it just re-runs \textbf{OO} at each timestamp, while its knowledge about the demand is still estimated from historical data, in the same way as \textbf{OO}. Comparing to the state-of-the-art \textbf{IO}, our approach \textbf{MANS} (with $w$=1) offers about 20\% improvement in \texttt{NV} and in \texttt{SC}. The gain comes from: i) we directly optimize \texttt{NV} in our reward structure, and ii) instead of relying on estimates from historical data which is essentially to optimize on a proxy task, we directly learn by trial-and-error on the target task using deep reinforcement learning.

\noindent \textbf{Deployment Behaviours under Different Rewards. }
In this set of experiments, we investigate the impact of the weight $w$ between the two terms in our reward structure, as discussed in Eq.~\eqref{eq:agent-r}. Here we vary $w$ in the range [9/1, 7/3, 1, 3/7, 1/9], which gives different importance to the two reward terms $g^{\texttt{\,SC}}$ and $g^{\texttt{\,PM}}$. We train our controller under different $w$, and report the performance of those variants in Fig.~\ref{fig:weights} as well as Table.~\ref{tbl:overall}. We see that as we decrease the weight, \texttt{NV} drops while \texttt{SC} increases. This is expected, as we essentially give more reward on \texttt{SC}, while ignoring actions that lead to high \texttt{NV}. On the other hand, we also see that \texttt{SC} is less sensitive to $w$ than \texttt{NV}. We see that the increase in \texttt{SC} is very small while $w$ decreases from 7/3 to 1/9. This is because that \texttt{SC} cares about the ubiquity of the service, and thus once the stations in the deployment plan can cover the critical mass of the candidate stations, \texttt{SC} tend to be stable. 
\rev{Fig.~{\ref{fig:30day-sc}} and Fig.{~\ref{fig:30day-nv}} shows the system performance under different $w$ as a function of the demand distribution over temporal domain (for 30 days). In particular, we show the 30 day trend of \texttt{SC} and \texttt{NV} for $w$ = [9/1, 1, 1/9]. We see that under different deployment plans \texttt{SC} remains relatively flat. However, the \texttt{NV} is more sensitive, as shown in Fig.{~\ref{fig:30day-nv}}. We see clearly that if we reward $g^{\texttt{\,PM}}$ heavily, the \texttt{NV} on certain days is much higher. It is possible that under such settings, our controller learns to deploy stations temporarily at hot spots on particular days, serving substantial amount of high-value orders. This may bring extra income but won't necessarily increase the service coverage, since the POIs may already be well covered by neighbouring stations.}

\noindent \textbf{Performance vs. Deployment Variability. }
Finally in this set of experiments, we evaluate the performance of our approach with respect to different level of deployment variability considered in our controller. As discussed in Sec.~\ref{sub:hierarchical}, our approach uses a hierarchical controller where the high-level controller generates a categorical \textit{deployment prior} ($S_t^{i+}$, $S_t^{i-}$), determining roughly how many new/existing stations should be deployed/removed for each category. Here we vary the magnitude of this deployment prior, and consider four \textit{action scales}: [0, 0.05, 0.1], [0, 0.1, 0.2], [0, 0.15, 0.3], and [0, 0.2, 0.4], each of which has three levels. Here 0 means there should be no stations to be deployed or removed. Therefore, the first scale [0, 0.05, 0.1] means that at each timestamp there could be no, 5\% or 10\% of the total stations removed or deployed. Intuitively, this scale specifies how much our controller can explore in the search space. For instance, if the highest scale [0, 0.2, 0.4] were selected, in the extreme cases, the controller can add 40\% of new stations, while removing 40\% of the existing ones in one timestamp. Fig.~\ref{fig:actions} shows the improved \texttt{SC} and \texttt{NV} of different variants of our controller compared with the baseline \textbf{HD}. We see that as we increase the scale, i.e., our controller has more room to explore, the \texttt{NV} is relatively stable, but drops a bit as the scale reaches the highest. This means our approach can robustly optimize towards the goal of \texttt{NV}, while at high scales the increased randomness in generated deployment plans may deteriorate \texttt{NV}, e.g., deploying more stations may cost too much. On the other hand, for \texttt{SC} we see that it generally increases, which is also expected, since if more stations are allowed to be altered, the controller is likely to propose plans that quickly fill up the candidate stations to improve the service coverage.


\section{Discussion}
\label{sec:discussion}

\noindent\rev{\textbf{Practicality.}
In practice, the cost of building mobility infrastructure in urban space (e.g., developing parking spaces) can be prohibitively expensive, where dynamical deployment of stations for shared e-mobility systems seems to be an impossible mission. However in the real world, instead of building/owning their own parking/charging infrastructure, those systems typically rent a certain number of parking spaces or charging docks in existing car parks, e.g., those operated in big shopping centres or transportation hubs. Therefore, the shared e-mobility systems can actually adjust their infrastructure in a flexible way without incurring significant overhead. For instance, if some charging stations / parking spaces are not needed during certain days, they will be ``returned to'' the car park providers, e.g., becoming available to the general public, and thus no extra cost would be imposed to the shared e-mobility systems. For the real-world shared e-mobility system studied in our paper, from the historical data we find that there are already stations which are only rented for a few days per week (e.g., weekends), while during the other days they are labelled as ``off-line'' in the app, and thus not accessible by the users. This confirms the potential and practicality of our approach, in that if we can discover deployment plans over space and time that reflect such knowledge, e.g., we only need certain stations on weekends, the shared mobility systems could better understand their needs and negotiate a better deal with the car park providers before deploying their infrastructure.}

\noindent \rev{\textbf{Collaborative vs. Competitive Agents.}}
\rev{
In typical MARL formulation, the behaviour of agents could be either collaborative or competitive. In our approach, we find that the agents tends to be collaborative, since in our case each agent manages the infrastructure deployment of a local region, while we explicitly allow them to observe both their local and global states. This enables the agents to interact with not just local environment, but also learn to better cooperate with the others that have connections with them. We observe that agents of neighboring regions, or those with high correlations in terms of user demand (i.e., users frequently travel from one region to another) tend to deploy stations together, or within close temporal proximity. This is also because our budget function is more tolerating, i.e., we only requires the deployment cost should not exceed the total income (GMV), encouraging the agents to deploy more stations in seek for better income and service coverage (as shown in Eq.{~\eqref{eq:agent-r}}). In cases where the budget is tight (e.g., more penalty in the reward function), the agents may also exhibit competitive behaviour, which would compete for limited resources to obtain more reward. 
}

\noindent \rev{\textbf{Limitations.}}
\rev{
Our approach uses reinforcement learning to search for better infrastructure deployment strategies, which essentially works in the way of trail-and-error. Therefore, unlike the exact or analytical optimization methods such as~{\cite{Du:KDD:2018}} and~{\cite{Liu:SIGSPATIAL:2019}}, our approach is not \textit{complete}, i.e., there is no theoretical guarantee that our approach will always converge to the optimum. However, the advantage of our approach is that it can capture the complex interactions between the users and the system in this infrastructure optimization problem, while the exact or analytical optimization approaches often need to make oversimplified assumptions to make the problem tractable. On the other hand, our approach uses sampling to propose and evaluate potential deployment plans, which is naturally less computationally efficient than the exact approaches that rely on heuristics. This could be a limitation for real-time optimisation, but in our context the deployment plans can be learned offline, i.e., the deployment plans could be pre-computed before actually implementing them, and therefore computational efficiency won't be a major limiting factor in practice. 
}



\section{Conclusion}
\label{sec:conclusion}

In this paper, we investigate the deployment optimization problem for shared e-mobility systems, aiming to find the optimal deployment plan with which the system can achieve the desired service coverage, net revenue, and demand satisfied rate while being profitable. We design a high-fidelity simulation environment to capture the key operational details the shared e-mobility systems at fine granularity, and calibrate the environment with rich data collected from a real-world shared e-mobility system over 12 months. To tackle the deployment optimization problem, we propose a novel multi-agent deep neural search algorithm, which employs hierarchical controllers to generate possible deployment plans. The controllers are trained using a multi-simulation paradigm within our simulation environment, which learns to propose better plans in the next iteration. The proposed approach has been evaluated extensively, and experimental results show that: i) our neural search algorithm significantly outperforms both the baseline and state-of-the-art optimization approaches, in various metrics including service coverage and net revenue; ii) by adjusting the weight between objectives, our search algorithm can optimize towards different directions, providing desired deployment plan for different cases; iii) the proposed hierarchical controller architecture reduces the search cost, where tuning the action space of the high-level controller has substantial impact on balancing exploration and exploitation.

\smallskip
\smallskip
\noindent \textbf{Acknowledgment} 
This work was supported in part by The Alan Turing Institute under the EPSRC grant EP/N510129/1.

\bibliographystyle{IEEEtran}
\bibliography{10-main}

\end{document}